\documentclass[a4paper]{article}

\usepackage{INTERSPEECH2022}
\usepackage[inline]{enumitem}
\usepackage{multirow}
\usepackage{tipa}
\usepackage{caption}
\usepackage{subcaption}

\title{Investigating the Impact of Cross-lingual Acoustic-Phonetic Similarities on Multilingual Speech Recognition}
\name{Muhammad Umar Farooq, Thomas Hain \thanks{This work was partly supported by LivePerson Inc. at the Liveperson Research Centre.}}
\address{Speech and Hearing Research Group, University of Sheffield, UK.}
\email{\{mufarooq1,t.hain\}@sheffield.ac.uk}

\begin{document}

\maketitle

\begin{abstract}
Multilingual automatic speech recognition (ASR) systems mostly benefit low resource languages but suffer degradation in performance across several languages relative to their monolingual counterparts.
Limited studies have focused on understanding the languages behaviour in the multilingual speech recognition setups.
In this paper, a novel data-driven approach is proposed to investigate the cross-lingual acoustic-phonetic similarities. 
This technique measures the similarities between posterior distributions from various monolingual acoustic models against a target speech signal. Deep neural networks are trained as \textit{mapping networks} to transform the distributions from different acoustic models into a directly comparable form. 
The analysis observes that the languages `closeness' can not be truly estimated by the volume of overlapping phonemes set. Entropy analysis of the proposed mapping networks exhibits that a language with lesser overlap can be more amenable to cross-lingual transfer, and hence more beneficial in the multilingual setup.
Finally, the proposed posterior transformation approach is leveraged to fuse monolingual models for a target language. A relative improvement of $\sim$8\% over monolingual counterpart is achieved.
\end{abstract}

\noindent\textbf{Index Terms}: automatic speech recognition, multilingual, acoustic-phonetic similarities, model fusion

\section{Introduction}
\label{sec:intro}

Multilingual automatic speech recognition (ASR) systems have attained significant attention over the past decade. Motivation for multilingual speech recognition includes;
\begin{enumerate*}[label=(\roman*)]
\item having a single unified model capable of recognising speech of diverse languages \cite{Pratap2020,hou20} and
\item using shared language representations to improve ASR performance in the low resource settings \cite{abate20,tachbelie20,imseng14,martin16,besacier14,vu13_interspeech,tong18,huang13}.
\end{enumerate*}
Previous works show that multilingual ASRs do not yield significant reduction in Word Error Rate (WER) for various languages including English, German, French and several others \cite{Pratap2020,hou20,conneau21}. Many open source data sets are available for these languages and are generally regarded as resource-rich languages. Most of the multilingual models, which include resource-rich languages, are trained on unbalanced data. So, degradation in the performance of these languages is attributed to increased confusion for them in the multilingual setup \cite{Pratap2020}. 
\textit{Conneau et al.} \cite{conneau21} mention language interference as the reason of the increased error rate.
However, no proof of these concepts could be found in the literature.

Earlier studies, in the context of efficient data sharing among languages for multilingual setup, are on Context Independent (CI) acoustic models \cite{kohler96}. Context Dependent (CD) acoustic-phonetic similarity was first studied by \textit{Imperl et al.} \cite{imperl00} and later extended by \textit{Le et al.} \cite{bac06}. Both of these studies measure distance between two polyphones as a weighted sum of monophonic distances of these polyphones. In \cite{bac06}, a knowledge based approach is applied which needs considerable manual efforts as it requires a hierarchical graph to measure the distance between monophones. Furthermore, the phonemes are clustered on IPA symbols. 


Recently, some efforts have been made to interpret the learning of multilingual speech recognition systems \cite{zelasko20,feng21}. Phoneme Error Rate (PER) of each phoneme in monolingual ASR was compared with that of multilingual system \cite{zelasko20}.
However, no monotonic trend was observed with the growing number of languages the phoneme shared. The authors described this as \textit{``unexpected"} because the phonemes shared by more languages provide more training data and thus the expected error trend would be decreasing.
Motivated by the fact that many languages with significant phonemes overlap pose performance degradation in the multilingual setups \cite{Pratap2020,hou20,conneau21,zelasko20,feng21}, the objective and contribution of this work is to study acoustic-phonetic similarities to understand if the cross-lingual phoneme sharing is truly a sharing and how does it impact on multilingual speech recognition setups?

To that end, a novel technique is proposed to estimate the cross-lingual acoustic-phonetic similarities for CD hybrid DNN-HMM acoustic models. Hybrid DNN-HMM system is preferred over end-to-end (e2e) modelling to avoid influence of the entangled language model in e2e speech recognition systems. An equal amount of three West Germanic languages (English, German and Dutch) is used to study the impact on multilingual performance. Behaviour of the monolingual acoustic models against a speech signal of the target language is studied by differentiating the posterior distributions. To compare distributions of source and target AMs, a separate regression neural network is trained for each \textit{$<$source, target$>$} pair to map posteriors from a source language AM to the posteriors of the target language AM.

\vspace{-0.5em}
\section{Cross-lingual acoustic-phonetic similarities}
\label{sec:amsim}
\vspace{-0.5em}
The motivation of research into multilingual speech recognition is based on an assumption that the articulatory representations of phonemes are very close across the languages and can be considered language independent units \cite{tanja01}. However, several languages with substantial cross-lingual phoneme sharing exhibit poorer performance in multilingual setups. This calls for a study to understand the reason of degradation or improvement in multilingual setups when compared with corresponding monolingual systems. 


Hybrid DNN-HMM systems yield better performance than the conventional GMM-HMM based ASRs and outperform e2e ASRs with the limited amounts of training data \cite{lfmmi}. Furthermore, the output from e2e speech recognition systems is influenced by the entangled language model which adversely effects the acoustic analysis.

In hybrid speech recognition systems, a deep neural network is trained to produce a posterior distribution of tied states of HMM models.
The total number of states (and thus output layer dimension of DNN) is reduced by clustering many polyphonemes together.
Each language yields a different phonetic decision tree in its monolingual ASR. Thus the number of tied states differs for each language and the posterior distributions are not directly comparable across the languages.

In this work a data-driven approach, to transform posteriors from diverse models to a directly comparable form, is proposed. A similarity measure is calculated between the posterior distributions from different models against a given speech signal to estimate the cross-lingual acoustic-phonetic similarities. 

\begin{figure}[tb]
\centering
\includegraphics[width=\linewidth]{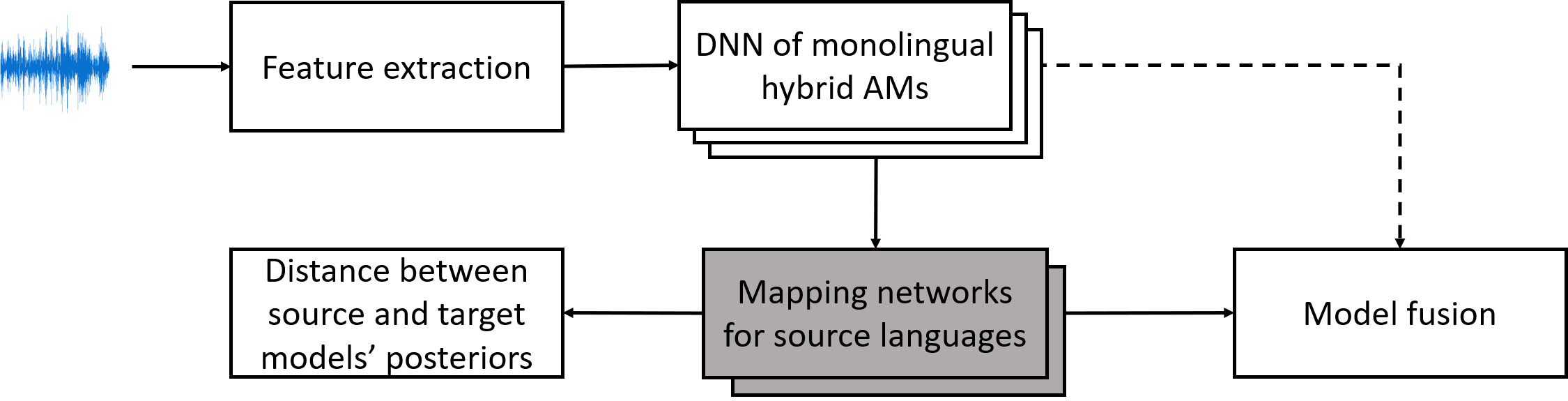}
\caption{Proposed system architecture}
\vspace{-0.5em}
\label{fig:archi}
\centering
\end{figure}

\vspace{-0.25em}
\subsection{Similarity measure}
\vspace{-0.25em}

Let $M_{A}$ and $M_{S_{i}}$ be the monolingual acoustic models of target and source languages respectively. The target language is the language for which the similarity is being measured against the source languages.
A regression neural network $N_{S_{i}A}$ is trained to translate posteriors $P_{S_{i}}$ of dimension $d_{S_{i}}$ from $M_{S_{i}}$ to the posteriors $P_{S_{i}A}$ of dimension $d_{A}$ where $d_{A}$ is the dimension of posteriors from $M_{A}$. An underlying assumption is that this \textit{mapping network} is able to learn some language related relationships between posterior distributions of source and the target acoustic models. For example, the network could learn the phonemes of target language which are more amenable to cross-lingual transfer than the others. A few hours of speech data can give thousands of examples that provide sufficient training data for \textit{mapping network}.
Kullback-Leibler (KL) divergence, the most widely used measure to differentiate two posterior distributions \cite{kldiv}, is calculated as a similarity measure between the posterior distributions from the target language AM and the mapped posterior distributions from the source language AM. The proposed system architecture is shown in Fig. \ref{fig:archi}.

Let \(X=\{x_{1},x_{2},\dotsc,x_{T}\}\) be a set of observations of target language, for which posterior distributions (\(P^{Z}=\{p_{1},p_{2},\dotsc,p_{T}\}\) where \(Z \in (A,S_{i}A) \)) are attained from all monolingual acoustic models. Posteriors from source acoustic models ($P^{S_{i}}$) are mapped to target posteriors ($P^{S_{i}A}$) using \textit{mapping network} $N_{S_{i}A}$. The similarity between a source and the target language for a given set of the observations is calculated as 
\vspace{-0.25em}
\begin{equation}
\label{eq:kl}
D_{X}(M_{T},M_{S_{i}})=\frac{\sum_{t=1}^{T} p^{A}_{t} \cdot (\log p^{A}_{t}- \log p^{S_{i}A}_{t})}{T}
\end{equation}

\section{Experimental Setup}
\label{sec:exp}

\vspace{-0.25em}
\subsection{Data set}
\vspace{-0.25em}

Experiment results are reported using three languages (English \textit{en}, German \textit{de} and Dutch \textit{nl}) of West Germanic family. From previous works, the performance of these languages is either degraded or show a very minor improvement in multilingual setups despite of a sufficient number of the shared phonemes. 
The sharing of cross-lingual phonemes is tabulated in Table \ref{tab:phShare}.

The experiments are carried out using portions of the Multilingual LibriSpeech (MLS) data set \cite{mls}. 
For training and evaluation of monolingual AMs, 30 hours and 2 hours are randomly sampled from corresponding sets of the MLS corpus respectively. Limited amount of data for each language is used in this study due to several reasons.

\begin{itemize}
 \item The amount of data is restricted as an ablation study to the argument that the performance of these languages degrades in the multilingual setups due to their very strong monolingual counterparts (trained on much more data being resource-rich languages) \cite{Pratap2020}. 
 \item On phonetic level, 30 hours are sufficient for acoustic-phonetic similarities analysis as they provide millions of examples for \textit{mapping network} training.
 \item Experiments with the limited data for model fusion, provides a realistic scenario for low resource languages and the technique can be extended for resource-deficient languages if the outcome is encouraging.
\end{itemize}

Baseline multilingual ASR is trained by mixing data from all the languages and hence trained on 90 hours of speech data. 
To train the \textit{mapping network}, the 30 hours train set is further divided into 29 hours of training and 1 hour for validation.

\begin{table}[]
 \centering
 \caption{\% Cross-lingual phoneme shares}
 \label{tab:phShare}
 \begin{tabular}{c|cccc}
 \hline
 \hline
 &\multicolumn{4}{c}{Shares of}\\
 \hline
 \parbox[t]{2mm}{\multirow{4}{*}{\rotatebox[origin=c]{90}{shared with}}}&&\textit{en}&\textit{de}&\textit{nl} \\
 &\textit{en}&100\%&70.58\%&69.48\%\\
 &\textit{de}&72.96\%&100\%&88.59\%\\
 &\textit{nl}&69.8\%&85.14\%&100\% \\
 \hline
 \hline
 \end{tabular}
\end{table}


%


\subsection{Similarity measure}
\vspace{-0.25em}
All acoustic models used for experiments in this work are trained using lattice-free MMI criterion (LF-MMI) \cite{lfmmi}. Though the study can easily be extended for polyphones of any context, left biphones are modelled for experimentation here and the term \textit{biphones} will be used during the remainder of the paper.

Given the partially overlapped phoneme sets of two languages (the target and a source language), there are the following subsets of biphones:

\begin{itemize}

\item \textit{Shared biphones:} Due to the overlapping cross-lingual phoneme set, there would be some biphones which occur in both (source and the target) languages.
\begin{itemize}
\item \textit{Shared seen biphones (SS):} The set of shared biphones which are seen by both languages during training of their acoustic models.
\item \textit{Shared unseen biphones (SU):} The set of shared biphones which are never seen by source language in its train set.
\end{itemize}
\item \textit{Unshared biphones (U):} The biphones of the target language that are never seen by source language due to non-overlapping phonemes.
\end{itemize}

\begin{table}[b]
\centering
\vspace{-1em}
\caption{Baseline ASR performance}
\vspace{-1em}
\label{tab:baseline}
 \begin{tabular}{c|ccc}
\hline
\hline
\multirow{2}{4em}{Language}&\multicolumn{3}{c}{WER/PER}\\
\cline{2-4}
&\textit{mono}&\textit{multi}&\textit{mono-lm}\\
\hline
English \textit{(en)}&43.84/28.10&47.48/31.06&46.40/31.06\\
German \textit{(de)}&37.77/26.86&40.81/28.35&38.11/28.35\\
Dutch \textit{(nl)}&37.94/21.40&58.84/36.16&52.33/36.16\\
\hline
\hline
\end{tabular}
\end{table}


A similarity measure is calculated for each of the aforementioned case and
an analysis of trends in distance measure is carried out to comprehend the language similarities and their behaviour in the multilingual setups.

\subsection{Model fusion}

The \textit{mapping network} is further used to map posterior distributions from several models to a uniform (target language) dimension. The weighted sum of these mapped posteriors is then used for ASR decoding. 

For a given observation at time $t$, the final posterior vector is given as;
\begin{equation}
\label{eq:wp}
p_{t}^{F}=w_{T} \cdot p_{t}^{A} + \sum_{i=1}^{N} w_{i} \cdot p_{t}^{S_{i}A}
\end{equation}
where $w$ are the scalar weights assigned to each posterior vector such that \(\sum w=1 \) and $N$ is the number of source languages.

\begin{table}[t]
\centering
\caption{Posterior distribution similarity for the \textit{en} test set}
\label{tab:en}
\vspace{-1em}
\begin{tabular}{p{0.05\linewidth}|p{0.12\linewidth}|p{0.17\linewidth}|p{0.20\linewidth}|p{0.20\linewidth}}
\hline
\hline
AM&Biphone subsets&\% Correct SAMC&KL-Div (SAMC)&Entropy (SAMC)\\
\hline
\hline
\multirow{4}{4em}{\textit{de}}&\textit{SS}&37.08 &1.32 (0.32)&2.43 (1.4) \\
&\textit{RSS}&13.30 &1.75 (0.84)&3.22 (2.64) \\
&\textit{RSU}&- &1.88&3.40 \\
&\textit{RU}& - &2.27&3.59 \\
\hline
\multirow{3}{4em}{\textit{nl}}&\textit{SS}&38.04&1.23 (0.32)&2.21 (1.29) \\
&\textit{RSS}&12.90 &1.64 (0.82)&2.95 (2.44) \\
&\textit{RSU}&- &1.71&2.86 \\
&\textit{RU}& - & 1.81&3.1\\
\hline
\hline
\end{tabular}
\end{table}

\section{Results and Discussion}
\label{sec:disc}

\vspace{-0.25em}
\subsection{Baseline multilingual ASR}
\vspace{-0.25em}
Monolingual (\textit{mono}) baseline systems are the language dependent acoustic, pronunciation and language models which are trained on a language specific data set. The train sets of all the languages are then mixed to train multilingual (\textit{multi}) acoustic and language models.
The multilingual acoustic model is then used with monolingual Language Model (LM) of the target language which is termed as \textit{mono-lm} in the reported results. The results of the baseline systems for all the languages are given in Table \ref{tab:baseline} in terms of WER and PER. Since only the language model is changed for \textit{mono-lm}, thus the PER remains unchanged when compared with that of multilingual system.
The results show that the error for all the languages increases in multilingual setup in spite of the balanced data duration for each language. This indicates that the reason of performance degradation in multilingual setups can not only be attributed to rich resources of these languages or unbalanced data sampling.

\begin{figure}
\centering
\includegraphics[width=\linewidth]{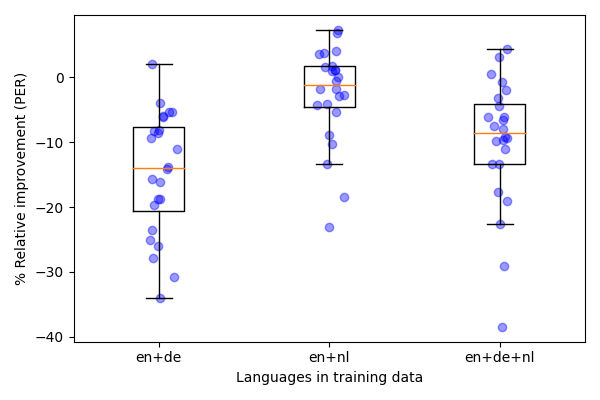}
\caption{\% Relative improvement in PER per shared phoneme compared with monolingual ASR for \textit{en} target language}
\vspace{-0.5em}
\label{fig:relPER}
\centering
\end{figure}

According to the assumption of multilingual systems \cite{tanja01} discussed earlier in section \ref{sec:amsim}, if the articulatory representations of phonemes are considered language independent units then the performance of shared phonemes should improve with the more training data in the multilingual systems. The languages, being studied here, have an overlapping set of 24 phonemes. As a case study of \textit{en} as the target language, the relative improvement in PER of shared phonemes is analysed with gradually increasing the languages in the training data. It is evident from the Fig. \ref{fig:relPER} that even the performance of shared phonemes is degraded in bilingual and multilingual setups, though \textit{nl} is less detrimental than \textit{de} for the \textit{en} language.
\begin{table}[b]
\centering
\caption{Posterior distribution similarity for the \textit{nl} test set}
\vspace{-1em}
\label{tab:nl}
\begin{tabular}{p{0.05\linewidth}|p{0.12\linewidth}|p{0.17\linewidth}|p{0.20\linewidth}|p{0.20\linewidth}}
\hline
\hline
AM&Biphones subset&\% Correct SAMC&KL-Div (SAMC)&Entropy (SAMC)\\
\hline
\hline
\multirow{4}{4em}{\textit{en}}&\textit{SS}&46.51&1.33 (0.37)& 1.75 (0.95)\\
&\textit{RSS}&15.43&1.87 (1.26)& 2.36 (1.77)\\
&\textit{RSU}&-&2.113&2.66 \\
&\textit{RU}&-&2.173&2.69\\
\hline
\multirow{3}{4em}{\textit{de}}&\textit{SS}&40.90&1.44 (0.46)&1.89 (1.19) \\
&\textit{RSS}&17.30& 1.88 (1.15)&2.33 (1.81) \\
&\textit{RSU}&-&2.04& 2.43 \\
&\textit{RU}&-&2.13&2.58\\
\hline
\hline
\end{tabular}
\end{table}

\begin{table}
\centering

\caption{Posterior distribution similarity for the \textit{de} test set}
\vspace{-1em}
\label{tab:de}
\begin{tabular}{p{0.05\linewidth}|p{0.12\linewidth}|p{0.17\linewidth}|p{0.20\linewidth}|p{0.20\linewidth}}
\hline
\hline
AM&Biphones subset&\% Correct SAMC&KL-Div (SAMC)&Entropy (SAMC)\\
\hline
\hline
\multirow{4}{4em}{\textit{en}}&\textit{SS}&43.56 & 0.83 (0.22)&1.76 (1.07)\\
&\textit{RSS}&14.90 & 1.21 (0.74)& 2.52 (2.12) \\
&\textit{RSU}&- &1.15&2.21 \\
&\textit{RU}& - &1.27&2.54\\
\hline
\multirow{3}{4em}{\textit{nl}}&\textit{SS}&36.37&1.05 (0.31)&1.95 (1.24) \\
&\textit{RSS}&13.68& 1.38 (0.82)& 2.5 (2.09)\\
&\textit{RSU}&- &1.54&2.67\\
&\textit{RU}& - &1.41&2.56\\
\hline
\hline
\end{tabular}
\vspace{-1em}
\end{table}

\vspace{-0.25em}
\subsection{Similarity analysis}
\label{sec:sa}
\vspace{-0.25em}




Table \ref{tab:en} shows the similarity measure for English (\textit{en}) test set when passed through German (\textit{de}) and Dutch (\textit{nl}) acoustic models. In clustering, shared seen biphones \textit{(SS)} may share the same cluster with unseen \textit{(SU)} or unshared biphones \textit{(U)} and vice versa. So for insightful observations, analysis is restricted to the clusters which have only one biphone (shown with \textit{``R"} prefix). It is clear from the results that KL divergence increases from shared seen biphones towards unshared biphones.

\begin{figure*}
\centering
 \begin{subfigure}[b]{\textwidth}
 \centering
 \includegraphics[width=\textwidth]{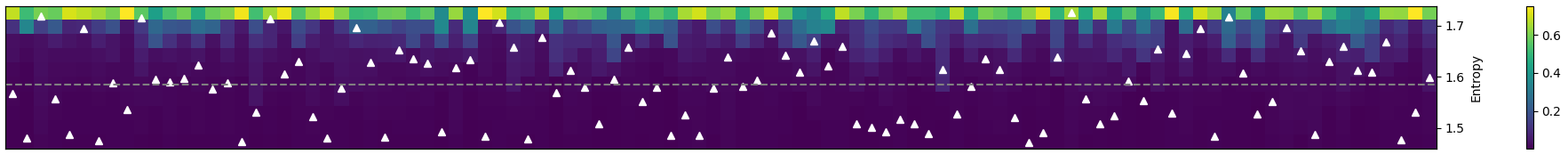}
 \vspace{-1em}
 \caption{de to en mapping network}
 \end{subfigure}
 \begin{subfigure}[b]{\textwidth}
 \centering
 \includegraphics[width=\textwidth]{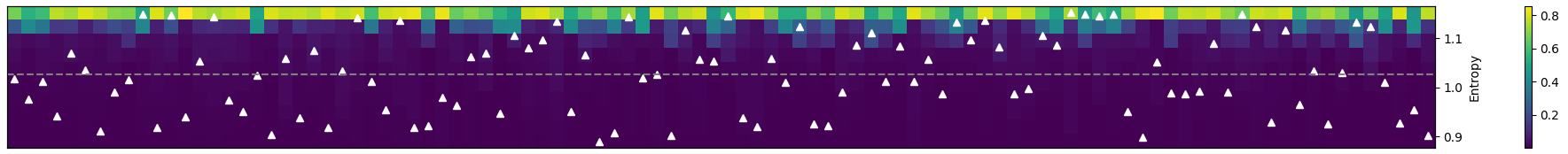}
 \caption{nl to en mapping network}
 \end{subfigure}
\captionsetup{singlelinecheck=off}
\caption{Posteriorgram and entropy plot of $N_{de-en}$ and $N_{nl-en}$ mapping networks with one-hot vectors as input. Behaviour for only hundred source biphone classes is shown. Posteriorgram shows sorted probabilities of top ten mapped classes. Each box on horizontal axis is a one-hot vector of source language and on vertical axis is the probability of a mapped output class}
\label{fig:entropy}
\end{figure*}



For shared seen biphone sets (\textit{SS} and \textit{RSS}), being present in both (source and the target) languages, mapping network should learn one-to-one mapping to same biphone class if the multilingual assumption holds true. This measure is calculated as percentage of these biphones recognised correctly by source acoustic model and reported as `\textit{correct Source Acoustic Model Class (SAMC)}' in the results. The results (low percentage of \textit{``Correct SAMC"} in tables) show that the source acoustic models are not very good at recognising these biphones but the lower values of KL divergence (\textit{KL-Div} in parenthesis) indicate that it is easier for \textit{mapping network} to learn a one-to-one mapping in these cases. It implies that the source ASR has a pattern in errors of these biphones sets which mapping network could learn easily. On analysis, it appears that the source acoustic models confuse these biphones with several close biphones. For example, a biphone \textbackslash\textipa{i}\textbackslash,\textbackslash\textipa{z}\textbackslash from English test set is frequently confused with \textbackslash\textipa{i}\textbackslash,\textbackslash\textipa{s}\textbackslash and \textbackslash\textipa{E}\textbackslash,\textbackslash\textipa{s}\textbackslash by German acoustic model. 

To study the confidence of the \textit{mapping network} in the mapped classes, it is sequentially fed the one-hot vectors as input. As a case study, the posteriorgram and entropy of \textit{mapping networks} from source languages to \textit{en} is analysed. Behaviour for the top \textit{n} source classes, which are mapped to the target classes with minimum entropy (more confidently), is observed. Entropy and posteriorgram from both networks, $N_{de-en}$ and $N_{nl-en}$, for same \textit{n} ($n=100$) is visualised in Fig. \ref{fig:entropy}. Only top ten most probable mapped output classes are shown in the sorted posteriorgram. It can be seen that for the same value of $n$, the entropy range is lower for \textit{nl-en mapping network} than the range for \textit{de-en} network. It evidences that \textit{nl-en mapping network} could learn better mappings and \textit{nl} phonemes are more amenable to transfer to \textit{en} phonemes. Furthermore, the entropy for each biphones subset is also measured which also shows the same trend as the KL divergence. It implies that the language similarities can be estimated through the entropy of the mapping network only without measuring KL-div between target and mapped posteriors.


\begin{table}[b]
\centering
\vspace{-1em}
\caption{Mean KL-divergence (as the cross-lingual similarity measure) and bilingual ASR performance}
\vspace{-1em}
\label{tab:avg}
\begin{tabular}{cccc}
\hline
\hline
\multirow{2}{4em}{Target Language}&\multicolumn{3}{c}{Source Languages(KL-Div/\% WER)}\\
\cline{2-4}
&en&de&nl\\
\cline{1-4}
en &0/43.84&1.56/46.35&\textbf{1.44}/44.46\\
de &\textbf{1.00}/38.94&0/37.77&1.18/39.46\\
nl &\textbf{1.60}/42.32&1.61/44.03&0/37.94\\
\hline
\hline
\end{tabular}
\vspace{-1em}
\end{table}

In the case of unseen \textit{(RSU)} and unshared \textit{(RU)} biphones, the source model tries to map them to the nearest biphone clusters. However, the performance of unshared biphones mapping goes further down in most of the cases.
Almost the same pattern is observed for the remaining two languages (Table \ref{tab:nl} and \ref{tab:de}). The average cross-lingual KL divergence along with the WER of bilingual ASRs is tabulated in Table \ref{tab:avg}. Diagonal entries are the monolingual ASRs. These results can be seen in comparison with cross-lingual phoneme sharing of Table \ref{tab:phShare}. For example, Table \ref{tab:phShare} infers that \textit{de} shares more phonemes with \textit{nl} compared to \textit{en}, but the mean KL divergence is smaller for \textit{en} (in Table \ref{tab:avg}). It implies that \textit{en} is more closer to \textit{de} than \textit{nl} and this claim is corroborated by \% WER of bilingual ASRs in Table \ref{tab:avg}. 
So, the phoneme sharing statistics are not a very informative metric to measure languages closeness and thus lead to degradation in the multilingual setups.


\begin{table}[t]
\centering
\vspace{-1em}
\caption{Performance of model fusion in \% WER}
\vspace{-1em}
\label{tab:res}
 \begin{tabular}{c|cccc}
\hline
\hline
Language&\textit{mono}&\textit{mono-lm}&\textit{mf}&Rel. imp\\
\hline
English \textit{(en)}&43.84&47.48&\textbf{43.43}&0.94\%\\
German \textit{(de)}&37.77&40.81&\textbf{36.16}&4.26\%\\
Dutch \textit{(nl)}&37.94&58.84&\textbf{34.94}&7.91\%\\
\hline
\hline
\end{tabular}
\vspace{-1em}
\end{table}


\vspace{-0.25em}
\subsection{Model fusion for multilingual ASR}
\vspace{-0.25em}
A multilingual acoustic model is imitated by fusing the target language and the mapped source language posteriors. The fusion is the linear weighted sum of all of these posterior distributions.
In Table \ref{tab:res}, the results of proposed model fusion (\textit{mf}) approach is compared with \textit{mono} and \textit{mono-lm} of Table \ref{tab:baseline}.
%
The setup of fusion technique makes results directly comparable to those of \textit{mono-lm} but the fusion outperforms even the \textit{mono} ASR. So, the reported relative improvement here is in comparison with the \textit{mono} ASR. However, fused posteriors give a relative gain of 6.4\% to 30\% when compared with \textit{mono-lm}.

\vspace{-0.25em}
\section{Conclusion}
\vspace{-0.25em}

In this work, cross-lingual acoustic-phonetic similarities are estimated by comparing the posterior distributions from source and the target acoustic models. A regression neural network is trained to map source languages posteriors to the target language posteriors. This study reveals that the behaviour of different languages for multilingual ASRs is more complex than predicting from cross-lingual phoneme sharing perspective. The languages which share more phonemes, does not guarantee performance gain in multilingual setups. The analysis observes that the phonemes with identical representations across languages are not acoustically identical.
Finally, the mapped posteriors are fused for decoding of a target language. A maximum gain of $\sim$8\% in relative improvement over the monolingual system is achieved. The relative gain increases to 30\% when compared with the fusion model's counterpart.


\bibliographystyle{IEEEtran}
\bibliography{refs}

\end{document}